# Identifying the Supply Chain of AI for Trustworthiness and Risk Management in Critical Applications

Raymond K. Sheh[1], Karen Geappen[2]

[1]Johns Hopkins University
[2]Independent Researcher
raymond@raymondsheh.org, karen@geappen.net

## Abstract

Risks associated with the use of AI, ranging from algorithmic bias to model hallucinations, have received much attention and extensive research across the AI community, from researchers to end-users. However, a gap exists in the systematic assessment of supply chain risks associated with the complex web of data sources, pre-trained models, agents, services, and other systems that contribute to the output of modern AI systems. This gap is particularly problematic when AI systems are used in critical applications, such as the food supply, healthcare, utilities, law, insurance, and transport.

We survey the current state of AI risk assessment and management, with a focus on the supply chain of AI and risks relating to the behavior and outputs of the AI system. We then present a proposed taxonomy specifically for categorizing AI supply chain entities. This taxonomy helps stakeholders, especially those without extensive AI expertise, to "consider the right questions" and systematically inventory dependencies across their organization's AI systems. Our contribution bridges a gap between the current state of AI governance and the urgent need for actionable risk assessment and management of AI use in critical applications.

## Introduction

The rise of modern AI systems, particularly those based on large, opaque models such as deep neural networks, presents significant challenges to the application of traditional concepts of trust, risk management, liability, and governance. There are increasing reports, both in industry publications and the popular media, of end-users with limited AI expertise running into trouble due to fundamental misunderstandings of how to risk-manage the behavior and outputs of AI systems. A topical example of a critical application is the legal profession. Despite the legal profession being at the heart of governance and risk management, Green, Sheh, and Heaney (2025) documented numerous recent cases where a lack of visibility into the origins of outputs resulted in significant problems in this domain.

The lack of understanding among many end-users and developers regarding how AI systems generate their outputs is exacerbated by the rapid increase in system complexity, distributed software development, and the growing use of AI-assisted coding by those without deep expertise in AI and software engineering. This increases the prevalence of obfuscated dependency webs and risk sources that interact unpredictably, making systems harder to comprehend, control, and maintain.

As the AI lifecycle grows ever faster, cutting-edge research becomes deployed, in formal and ad-hoc manners, in critical applications such as the food supply, healthcare, utilities, law, insurance, and transport, on a rapid timescale. In such a climate, researchers must be increasingly mindful of the crucial role they play in the supply chain of AI systems. They may be the only ones in a position to understand the risks associated with their selection of upstream entities and to clearly communicate this to those downstream and, ultimately, to the end-users.

We focus on AI systems that rely on machine-learned, high-dimensional statistical models, such as deep neural networks, to make decisions and generate outputs. This covers AI systems that include other capabilities such as retrieval augmentation and agentic functions. Critical applications are those where an incorrect decision can have a significant adverse impact on public health, life, and/or property. Examples include scheduling aircraft maintenance, setting legal precedents, routing emergency services, and developing new medications.

These critical applications are increasingly adopting AI for both generic tasks (with or without domain-specific modifications), such as document and meeting summarization, as well as for specialized tasks, including anomaly detection, data processing to support decision making, and code generation tailored to a specific application. In some cases this is done in an ad-hoc manner, without proper governance. For example, individual users may choose to use AI systems to summarize work emails or meetings without the approval or visibility of their organization.

Indeed, AI is sometimes used without even the end-user realizing it. For example, an update to a tablet running medical recording software might include an AI-enhanced predictive text keyboard. A user who is accustomed to the keyboard behaving in a certain manner when filling in patient records may not immediately realize that their inputs are now being augmented.

More general industry best practice for critical applications includes a comprehensive and rigorous supply chain risk management and governance strategy. This encompasses physical aspects, including ingredients, parts, and equipment. This may also include any software present, which incorporates code and sub-component provenance and traceability, to assess the product's risk.

Similar approaches to ensuring supply chain traceability for AI systems are still under development. Although we survey some salient examples in this paper, the governance requirements surrounding them and the translation of those requirements into technically actionable policy remain unclear. As was the case prior to the establishment of rigorous supply chain governance in other fields, end-users tend to be the most exposed to risk, liability, and marketing, despite being the least informed or empowered to take on and control this risk.

Supply chain visibility is a key gap in establishing risk management controls and improving trustworthiness in AI systems. In this context, we consider the supply chain to include all entities that contribute to the final output of the AI system.

We survey the state of AI supply chain risk management and then describe our proposed lightweight taxonomy, which improves the ability of all stakeholders across the AI lifecycle, including researchers, developers, integrators, and end-users, to identify the entities they depend on. Our contribution supports preliminary risk management, fosters communication among stakeholders, and lays the groundwork for the application of more formal AI risk management controls and frameworks as they mature.

## A Traditional Supply Chain Example

In traditional manufacturing, supply chain identification is vital to assessing risk and engendering public trust and acceptance. Establishing the provenance of each input to the end product and the party responsible for actions in the manufacturing process is also crucial to determining liabilities and any applicable regulations.

For instance, processed foods are typically marked with the manufacturer's name, product name, and batch number or expiration date. If a contaminated product is found, the source of the product and the activities performed on that product can be identified. Harm minimization and rectification can be undertaken, such as a product recall, while unaffected products can also be identified, further promoting trust.

A vital part of this trust and risk management is the recursive nature of this process. The manufacturer is obliged to keep track of the suppliers, ingredient names, and batch numbers of the inputs to their product. In some cases, this also extends to the machinery used to make the product. Recording the provenance of ingredients (and machinery) enables upstream suppliers to be informed if contamination is discovered. That upstream supplier can then notify other manufacturers who were supplied the same ingredient, from the same batch, to issue their recalls.

Rigorous supply chain governance also clearly defines liability for activities in the manufacturing process, such as transport or quality control. Identifying these entities further protects stakeholders by appropriately allocating liability to the different entities in the supply chain.

Traditional supply chains also provide end-users with supply chain information at different levels of detail and abstraction. For example, there is a standard taxonomy for recursively identifying characteristics of entities in the food supply chain by independently certified, end-user-facing attributes, such as "nut-free facility", "certified organic," or "FDA (Food and Drug Administration) approved." While imperfect, this practice helps to inform end-users with an appropriate level of detail, enabling them to make risk management decisions specific to their situation and risk acceptance level.

## AI Risk Management

The US National Institute of Standards and Technology (NIST) defines risk management as "the process of identifying, assessing, and responding to risk" (NIST 2018), while risk itself is defined as "a measure of the extent to which an entity is threatened by a potential circumstance or event." NIST expands on the concept of risk relating to AI systems through its AI Risk Management Framework (AI RMF) (NIST 2023). This framework provides a high-level taxonomy of AI risk, arranged in four main functions. Map relates to the identification of the overall context in which the AI system is used and associated risks. Measure relates to the measurement and analysis of those risks. Manage relates to the steps taken to address those risks. An overarching function of Govern addresses risks and measures associated with management and governance. The AI RMF primarily focuses on risk management for a single entity or stakeholder utilizing a single system, and mentions supply chains in passing within the Governance function.

The AI RMF is intended to be extended using domain, community, and organization-specific profiles. One example is the AI Risk-Management Standards Profile for

General-Purpose AI (GPAI) and Foundational Models (Barrett et. al. 2025). This profile places more emphasis on the supply chain, with a mention of the need to evaluate and document "... secure development and supply chain practices …" as part of the Measure function.

The AI RMF and their profiles are vital frameworks for managing risk, and they must be supported by a taxonomy of the entities in the supply chain that expose stakeholders to risk, so that these frameworks can be applied comprehensively and without blind spots.

As another example, the approach to AI risk management in Australia is described in the Australian Voluntary AI Safety Standard (Commonwealth of Australia 2024), which aligns with ISO/IEC 42001:2023 (International Organization for Standardization 2023) and the aforementioned NIST AI RMF. The Standard is comprised of 10 guardrails, the second of which is to "Establish and implement a risk management process to identify and mitigate risks". Within this guardrail, there are two activities: 2.1 - AI risk and impact management processes, and 2.2 - System risk and impact assessment.

As the standard is intended to be voluntary and use the organization's existing frameworks and processes, it does not indicate specifics on AI risk assessment; rather, it provides considerations. Examples include creating organizational risk tolerance for AI systems and identifying unacceptable AI risks. These can then be added to an organization's existing risk framework and risk processes. The standard provides guidance on AI-specific considerations, such as bias, potential harms, and, where appropriate, calls out guidance for working with AI suppliers and for procurement.

Focusing on the supply chain, the first step in managing risk is identification, which is where supply chain visibility is crucial. Much of the literature takes a narrow view, focusing on the provenance of the AI system, including the software, training data, and model. We expand on this and also consider the risks and supply chain entities associated with the normal development and use of these systems, as well as the risks caused by nefarious actors, whether inside or outside the organizations represented in the supply chain. We extend the visibility of the supply chain of AI to include all entities that influence the final output of the AI system, including those that host models, provide services, and aggregate data for use in both training and at runtime.

In focusing on the supply chain, we defer discussion of risks associated with, for example, bias or algorithm choice, to efforts such as the aforementioned frameworks, except where they directly relate to the supply chain, such as a lack of supply chain visibility that obfuscates potential bias.

## A Whole AI System Example

In a hypothetical large language model (LLM) based meeting summarization system, we might consider the following "ingredients" in its supply chain:

- The foundational model, including its training data, the implementation of the training system, and its hyperparameters.
- Any inputs into its fine-tuning and other processes or models that specialized it for a given domain.
- The sources of information that it might have retrieved, be it internal or external, via retrieval-augmented or other agentic behavior.
- System prompts and other contributions to its context.
- The implementation of the software system that the user uses to interact with the LLM.

Awareness of the "ingredients'" provenance enables us to consider the likelihood and consequence of "circumstances or events" that would cause the AI system to perform contrary to the intent, such as:

- The omission of important and contextual information.
- The misinterpretation and subsequent missummarization of information, such as domain-specific terminology.
- The generation of outputs that are unsupported by data, such as harmful hallucinations and misleading extrapolations.
- The unauthorized release of content, metadata, or logs.
- The system becoming silently degraded or unavailable.

These events pose significant threats in critical applications, such as test results being hallucinated in the summary of a meeting between a doctor and patient. Visibility into the supply chain might highlight contributors to such risks, such as identifying models in the supply chain with inappropriate guardrails or other checks, or datasets that include inaccurate medical information, such as outdated textbooks or medical-themed works of fiction.

## Existing AI Supply Chain Efforts

The problem of software supply chain risk management remains the subject of considerable research, despite its long history, in cybersecurity and beyond (NIST 2022). In recent years, this has been exacerbated by the rapidly expanding ecosystem of dependencies.

Projects like the Linux Foundation's System Package Data Exchange (SPDX) (Linux Foundation 2025) and the Open Worldwide Application Security Project (OWASP) Foundation's CycloneDX (OWASP Foundation 2025) have been undertaking efforts to standardize the definition, interpretation, and use of data fields to describe software modules, components, libraries, systems, services, and other resources. Such efforts are best known for their implications in the cybersecurity aspects of the software supply chain. This reflects the increasing prevalence and severity of supply chain attacks, as well as more recent geopolitical concerns, particularly given the growing complexity of software and its supply chains.

In recent years, both SPDX and CycloneDX have added fields relating to machine-learned models to document training datasets, provenance, metadata, licensing, and integrity. As with software, their extensions for AI models focus on the "ingredients" and associated information, such as licensing, maintenance, integrity, and security. Factors such as performance and intended use are not the primary focus of these efforts.

Model Cards (Mitchell et al. 2019) aim to document machine-learned models, focusing on their performance characteristics. Their original goal was to "... clarify intended use-cases … and minimize their usage in contexts for which they are not well suited …." Using the food analogy, AI model cards are more akin to a nutrition label, rather than an ingredients list. In practice, some model cards do include some documentation about the training data. Examples include those for GPT-3 (OpenAI 2025) and Salesforce for several of their AI products (Baxter and Schlesinger 2020).

On the governance side, efforts have been underway to establish standards for AI Bills of Materials (BOMs). The US Department of Homeland Security (DHS) Cybersecurity and Infrastructure Security Agency (CISA) Software Bill of Materials (SBOM) for AI Tiger Team initiative is a community effort that builds on existing AI documentation efforts, such as SPDX, CycloneDX, and Model Cards (CISA SBOM for AI Tiger Team 2025). It aims to identify fields that may require alternative or expanded interpretation for AI systems, and define new fields. It focuses on the model and training data, rather than algorithmic or implementation aspects.

The UK-based Trustable AI Bill of Materials (TAIBOM) project is a collaborative effort that takes a much broader view and seeks to cover the whole AI supply chain, "... from training data through the results that AI systems produce" (Techworks 2025). Their goal is to address the challenge of "... providing formal descriptions of AI systems and dependencies …" and of describing and validating subjective claims about the qualities of these systems. An ongoing project, at the time of writing it only considers a "... single instance neural network and the complex dependencies that lie within …." It aims to eventually consider much larger, multi-agent systems and integrate with efforts like those mentioned above.

The Australian Government, through its Australian Government AI technical standard (Commonwealth of Australia 2025), includes criteria relating to supply chain issues, framed from the perspective of the data supply chain. Of relevance to the management of risk associated with the resulting outputs of the AI system, Statement 13, "Establish data supply chain management processes" includes criteria for data that relate to use, fit-for-purpose, cybersecurity, quality, availability, and traceability. Similar to our approach, this standard considers "the flow of data across the AI system" beyond that used just for training.

Although much lighter in weight, our taxonomy, much like the TAIBOM project, covers a wide range of contributions to AI system outputs. It provides immediately actionable intelligence, even for less AI-proficient stakeholders, by identifying "who to ask" for detailed risk information. It similarly identifies where to apply existing and emerging frameworks such as Model Cards and SPDX/CycloneDX BOMs while enabling systematic identification of gaps in current risk visibility.

In terms of supply chain analysis, our work also resembles that of Gambacorta and Shretti (2025), who categorize AI supply chain entities into hardware, cloud, data, models, and applications, and discuss each entity's market share and macro-level risks. For example, they highlight that there are few alternatives to Nvidia, due to the entrenchment of their parallel computing platform, CUDA, in the rest of the AI supply chain. Furthermore, Nvidia is reliant on the Taiwan Semiconductor Manufacturing Company (TSMC), which poses a risk due to its reliance on a single manufacturer that is also potentially subject to geopolitical risks. This process is part of cybersecurity supply chain risk management best practice in identifying 'single point of failure' suppliers with reduced alternatives.

Where Gambacorta and Shretti (2025) focus on business and macro-level risks, our work targets technical sources of risks associated with each component and their effects on the AI system's outputs and fit-for-purpose. This reflects the novel aspects of AI supply chain risks not covered by traditional and industry best practices. Where non-AI systems may aggregate supply chain risk, an AI system may have individual supplier risk interact and influence the final AI system in a manner not easily deduced. By illuminating AI-specific aspects of the supply chain, we facilitate the development of effective governance controls to improve visibility into, and identification of, the risks associated with the AI system's implementation, procurement, deployment, use, maintenance, and retirement.

## AI Supply Chain Components and Entities

We divide the supply chain of AI into four components: **data**, **models**, **programs**, and **infrastructure**, each with several roles. Roles and entities may overlap in both directions. Entities may assume multiple roles across components, and multiple entities may contribute to a single role. The level of abstraction has been chosen, based on the aforementioned literature, to strike a balance between the level of detail that is useful to an end-user in applying other risk management controls and frameworks, such as those mentioned above. The roles that entities may assume in our taxonomy are:
- **Data: users**, **aggregators**, **hosts**, and **creators**.
- **Model: users**, **hosts**, and **creators**.
- **Program: users**, **hosts**, **integrators**, and **developers**.
- **Infrastructure: users**, **hosts**, **integrators**, and **developers**.

### Data

Many entities control data in the AI supply chain. We adopt a broad definition of data to include all information that ultimately results in the output of the AI system. Our definition covers instance-based datasets of training data, data generated through refinement processes such as reinforcement learning with human feedback (RLHF), symbolic background data such as formulas and rules, corpora of data (including those from the end-user) that might be referenced by retrieval-augmented systems, and sources of data on the Internet that may be accessed by agentic systems.

Note that our definition of data contains a nuance when it comes to the way that many current LLM systems are structured. Conventionally, end-user direct instructions, such as "write me a poem," are not part of the supply chain. However, an LLM prompt may contain a combination of direct instructions as well as other information that the end-user adds as context. For example, the end-user might provide a prompt of "write me a poem in a style similar to these other poems" and then append a selection of other poems to the prompt. This appended information, the selection of poems in our example, would be considered data in our discussion, even if the end-user's instruction itself is not.

We categorize the entities controlling data into the following roles. In AI systems developed internally to an organization using a well-defined corpus of training data, one entity may fulfill all of these roles. In contrast, for AI systems that draw their training data from numerous disparate sources, often scraped from across the Internet (so-called "web-scale" datasets), some roles may be fulfilled by multiple entities.

- **Data users** select and use the collections of data, such as those who use datasets of training data to train and fine-tune models.
- **Data aggregators** define collections of data (but do not necessarily store the actual data).
- **Data hosts** store data and make it available, without necessarily knowing that their data is part of a collection or dataset.
- **Data creators** create, label, and/or record the data in the first place.

This division highlights specific risks. For example, traditionally, an entity might create an image dataset by collecting images, collating them in some structured manner, and labelling them, before hosting them on a website. This entity is the **data creator**, **host**, and **aggregator**. The provenance of the data and entity responsible for actions on the data is clear and known.

For web-scale datasets, as explained by Carlini et. al. (2024), there is a split where the **data aggregator** merely links to **data hosts**. In this context, data hosts are entities that host user-generated content, such as social media platforms, code repositories, academic paper archives, and community-editable encyclopedias. That content is, in turn, created by **data creators**, such as contributors to a code base or online encyclopedia entry. Neither the **data hosts** nor **data creators** may be aware that their data has been aggregated in this manner, nor may they have necessarily consented to this aggregation.

**Data users** are exposed to risk due to a lack of visibility into changes made by **data hosts** and **creators**, such as the natural replacement or disappearance of web pages. There may also be malicious action, such as an attacker finding **data hosts** that they can control, including links to expired domains and social media accounts. By re-registering those domains or accounts, attackers have the opportunity to become malicious **data creators** the next time the dataset is updated or used. Carlini et. al. (2024) calls this split-view data poisoning.

Furthermore, if the **data aggregator** implements only limited curation and quality control, a malicious actor could easily masquerade as a legitimate **data creator** and have their content incorporated into such datasets. Carlini et. al. (2024) refers to this as frontrunning data poisoning.

Supply chain visibility enables data users to understand the creation and actions involved in creating a dataset, thereby improving fit-for-purpose and acceptable risk determinations. For example, a dataset with unauthenticated or otherwise unknown **data creators** may represent an acceptable risk when training a model to understand natural language but be unsuited for learning specialized terms or critical tasks.

Identifying (or verifying the lack of) **data creators** or **data aggregators** that are themselves AI systems is vital to controlling risks associated with the model ingesting synthetic data (Shumailov et. al. 2024). Where synthetic data is deliberately included, such as for masking private data, tracing the provenance of that data, its deviations from reality, and the expected impact on the model becomes important. In contrast, it is vital to understand if synthetic data is accidentally incorporated into a dataset, such as by scraping AI-generated articles from the Internet.

## Models

While Gambacorta and Shretti (2025) focus on foundational models, our taxonomy extends to all models within an AI system. In the context of machine learning, "model" usually refers to just the parameterized statistical model that encapsulates the training data, interpreted through the learning technique's inductive (or learning) bias (Mitchell 1997).

However, the popular media, vendors of AI systems, and government agencies tend to use the term "model" in a broader context, perhaps in acknowledgement of the fact that it can be impossible for someone outside of the model vendor to tell if a "model" is monolithic and machine-learned, versus a subsystem that consists of many parts, some machine-learned and others manually programmed.

For example, Nvidia defines a "model" as "... an expression of an algorithm that combs through mountains of data to find patterns or make predictions." (Parsons 2021). IBM defines a "model" as "... a program that has been trained on a set of data to recognize certain patterns or make certain decisions without further human intervention …" and that models "… apply different algorithms to relevant data inputs to achieve the tasks, or output, they've been programmed for." (IBM 2025). NIST Special Publication 800-218A on Secure Software Development Practices for Generative AI and Dual-Use Foundation Models defines an AI model as "… a component of an information system that implements AI technology and uses computational, statistical, or machine-learning techniques to produce outputs from a given set of inputs." (Booth et. al. 2024).

The Organisation for Economic Co-operation and Development (OECD) adopts a similarly broad definition, stating that a "model" is "… a core component of an AI system used to make inferences from inputs to produce outputs." and "… can be built manually by human programmers or automatically through, for example, unsupervised, supervised, or reinforcement machine learning techniques" (Organisation for Economic Co-operation and Development 2024).

To maintain compatibility with a broader audience outside of the machine learning community, we therefore adopt this broader definition. We define "model" to refer to not just the aforementioned statistical model, but also the components that the model creator considers part of the "self-contained" subsystem. These components may include symbolic guardrails and other rules and sub-models that are intended to transparently accompany the model wherever it is used, but are not directly exposed to users. In contrast, we consider the components that the model creator intends others to modify, such as system prompts, agents, and retrieval capabilities, as separate from the model.

Among the model types receiving much attention at the time of writing are foundational models. These are typically trained on massive amounts of data and offer generic capabilities, such as language understanding. These may be fine-tuned by additional models, such as low-rank adaptation (LoRA) models, which are trained on smaller quantities of application-specific data. Alternatively, specialized (non-foundational) models may be built from scratch for specific tasks.

Roles that entities in the **model** component of the supply chain may assume include the following:
- **Model users** select, combine, and use the models.
- **Model hosts** host the models and make them available, such as via an API.
- **Model creators** create, optimize, and maintain models. They utilize information from various sources, including datasets, symbolic rules, and human feedback.

## Programs

Each stage of the AI pipeline involves programs of some sort, be it to collate the data, create the models, run inference, or interface with the user. Gambacorta and Shretti (2025) highlight user-facing AI applications, and we extend this to all programs in the supply chain.

We define programs as any code, running anywhere in the AI supply chain, that is directly related to the AI system and is engineered in some way rather than learned. The distinction between "program" and "model" blurs in the current climate of AI-assisted code generation and "vibe coding." Programs include end-user AI applications, components of agents that are not learned, retrieval systems that augment context, and programs that train, fine-tune, test, and host models. We generally consider programmatic guardrails to be part of the **model** unless the **model creator** obtained them from a separate entity. Roles within the **program** component include:
- **Program users** specify, select, and use the programs, whether to train models or as end-users of AI applications. Users control specific program settings, such as hyperparameters in training.
- **Program hosts** make programs available to program users, such as via a web app or API.
- **Program integrators** set up programs on hosts. Integrators may also control specific program settings, such as system prompts.
- **Program developers** develop the underlying programs.

The mapping between Program and Model roles is not necessarily straightforward. For example, an AI chatbot application may allow a **program user** to select between multiple models. The role of the **model user** is divided between the **program user**, the **program developer**, and potentially the **program integrator**. For simple systems, the same entity may be simultaneously a **program user**, **program host**, and **program integrator**.

## Infrastructure

We combine the hardware and cloud computing components of Gambacorta and Shretti's (2025) AI supply chain into infrastructure. We define infrastructure as the hardware, networking, storage, cybersecurity, and other systems that support the data, models, and programs for training, interaction with the end-user, and otherwise.

Roles within **infrastructure** include:
- **Infrastructure users** directly select and use the infrastructure, be it to deploy programs, make models available, host data, or otherwise.
- **Infrastructure hosts** make the infrastructure available, be it internally or publicly.
- **Infrastructure integrators** set up host infrastructure.
- **Infrastructure developers** create hardware, operating systems, software platforms (e.g., CUDA), networking, storage, security, monitoring, and other host systems.

This breakdown of roles provides visibility into a wide variety of AI system structures, extending beyond just client and cloud. For example, an **infrastructure user** may be their own **infrastructure host** with a self-contained, on-premises AI appliance, set up by an **infrastructure**

**integrator,** and engage with several **infrastructure developers**.

## Illuminating the AI Supply Chain

A consumer seeking organic ice cream does not themselves trace the pesticide used by the farmer who grew the hay that fed the cow who made the milk for their ice cream. Similarly, our taxonomy should be used at an appropriate level of depth and abstraction to help end-users identify and apply risk-management controls relevant to their intended use of the AI system, their level of risk acceptance, and other precautions they may be taking.

We use the example of a hospital or hospital system, negotiating a contract for the implementation of a chatbot to advise the hospital's doctors. We highlight a few key entities and mention some basic risk management steps that may be relevant and yet non-obvious to the hospital. Naturally, this is a precursor and roadmap for applying a rigorous risk-management framework to the identified entities, such as the AI RMF, with an appropriate profile.

While the hospital is likely to be dealing directly with the **program integrator** who configures the system, they should also consider the **program developers** and **program hosts** who manage the provided service and include them in their risk assessment, using existing frameworks for cybersecurity, availability, redundancy, privacy, and other relevant factors.

Knowledge of the identities of the **model creators** is also vital. While most creators and vendors of foundational models have adopted a black box "trust us" approach and allow little to no tracing up their supply chain, at least knowing the identity and the version of model being supplied, along with their model cards and BOMs, allows the hospital to monitor widely used models for announcements of problems and concerns, particularly as they relate to their specific application.

Now consider that the chatbot has agentic capabilities, which the **program integrator** configures to access the hospital's own live information, such as admissions, records, and so on. The **program integrator**, as a **data user**, is responsible for configuring the chatbot to access the correct data. The hospital becomes a **data creator** and potentially a **data host** unless that is outsourced. This example highlights a frequently overlooked risk of such systems: they often require cleaner data than humans do.

Agentic systems are particularly at risk of being misled if the **data host** does not make contextual information obvious to the agentic system through their API, if the **program developer** of the agentic system did not consider that context, or if the **program integrator** did not provide a pathway for this context. As part of managing this risk, the hospital system, as **data creator** and **program user**, may identify "corner cases" in datasets for additional testing by the **data host**, **program integrator**, and **program developer**. They may also require the **program integrator** to disallow any agentic behavior that consults **data hosts** within or outside of a given list or that do or do not satisfy particular requirements. This would be like requiring an "all ingredients sourced from this region," "certified organic," or "made in a nut-free facility" label on a food product. If the final AI system leverages multiple other systems, the **program integrator** must understand any risk interactions between such systems.

As our final example, **program developers** and **integrators** may utilize both foundational and domain-specific models, with different **model creators** and **model hosts**. For example, the system may ordinarily use a medical-specific model to interpret certain inputs and silently fall back to a generic model if the medical-specific one is unavailable. As part of applying their chosen risk management framework, the hospital, on realizing that multiple models are in play, may forbid the chatbot from falling back silently. Instead, they may require the **program developer** or **integrator** to configure the chatbot to clearly indicate when it is running without the benefit of the medical-specific model. This way, the end-users are aware of the difference in the chatbot's behavior and capabilities, and act accordingly.

## Conclusion

We have presented a lightweight taxonomy for entities in the supply chain of AI that provides immediate, actionable intelligence, particularly for non-AI-proficient stakeholders. This is particularly important, given that AI introduces complexities not typically found in software supply chain risk management. By filling a critical gap and providing visibility into the web of AI system dependencies, our taxonomy enables these stakeholders to deploy current and emerging frameworks, as well as other governance controls, to manage risks associated with their use of AI systems.

As AI systems and their development cycles become increasingly complex and rapid, it becomes increasingly important for researchers, developers, and other contributors to the AI supply chain to be aware of and facilitate this visibility in support of the responsible deployment of their systems. Future work will include evaluating the applicability and effectiveness of this taxonomy in facilitating the end-user adoption of various AI governance controls and frameworks as they mature.